\documentclass[10pt,conference]{IEEEtran}
\IEEEoverridecommandlockouts
\usepackage{cite}
\usepackage{amsmath,amssymb,amsfonts}
\usepackage{algorithmic}
\usepackage{graphicx}
\usepackage{textcomp}
\usepackage{makecell}
\usepackage{hyperref}
\usepackage{subcaption}
\usepackage{microtype}
\usepackage{color, colortbl}
\usepackage{xcolor}
\def\BibTeX{{\rm B\kern-.05em{\sc i\kern-.025em b}\kern-.08em
    T\kern-.1667em\lower.7ex\hbox{E}\kern-.125emX}}

\definecolor{LightBlue}{RGB}{212, 250, 252} 

\begin{document}

\title{Domain-Specialized Object Detection via Model-Level Mixtures of Experts}

\author{\IEEEauthorblockN{Svetlana Pavlitska$^{1,2}$, 
Malte Stüven$^{2}$, Beyza Keskin$^{2}$,
and J.~Marius Zöllner$^{1,2}$}
\IEEEauthorblockA{
\textit{$^{1}$FZI Research Center for Information Technology} \textit{$^{2}$Karlsruhe Institute of Technology (KIT)}\\
pavlitska@fzi.de}
}


\maketitle

\begin{abstract}
Mixture-of-Experts (MoE) models provide a structured approach to combining specialized neural networks and offer greater interpretability than conventional ensembles. While MoEs have been successfully applied to image classification and semantic segmentation, their use in object detection remains limited due to challenges in merging dense and structured predictions. In this work, we investigate model-level mixtures of object detectors and analyze their suitability for improving performance and interpretability in object detection. We propose an MoE architecture that combines YOLO-based detectors trained on semantically disjoint data subsets, with a learned gating network that dynamically weights expert contributions. We study different strategies for fusing detection outputs and for training the gating mechanism, including balancing losses to prevent expert collapse. Experiments on the BDD100K dataset demonstrate that the proposed MoE consistently outperforms standard ensemble approaches and provides insights into expert specialization across domains, highlighting model-level MoEs as a viable alternative to traditional ensembling for object detection. Our code is available at \url{https://github.com/KASTEL-MobilityLab/mixtures-of-experts/}
\end{abstract}


\section{Introduction}
\label{sec:intro}
Object detection systems are increasingly deployed in safety-critical and real-world settings where both predictive accuracy and interpretability are essential. Ensemble methods are used to improve detection performance. However, in standard object detector ensembles where predictions from all models are simply concatenated and a suppression-based postprocessing step is applied, interpretability is inherently limited. The final decision emerges from heuristic postprocessing rather than from an explicit model-level mechanism that explains how individual detectors contribute. Once predictions are merged, it is difficult to attribute a retained detection to a specific expert, to understand why one expert dominated over another, or to analyze how expert behavior varies across input conditions. Non Maximum Suppression (NMS) further obscures attribution by discarding overlapping predictions without preserving information about suppressed alternatives or their originating models. Ensembles thus primarily improve performance but provide little insight into model agreement, disagreement, or domain-specific specialization.

Mixture-of-experts (MoE)~\cite{jacobs1991adaptive} models offer a principled alternative by combining multiple specialized networks through an explicit routing mechanism that enables both performance gains and interpretability through expert selection.

Recent work has made substantial progress on layer-level MoE architectures for transformer models\cite{fedus2021switch,dai2024deepseekmoe,riquelme2021scaling,lepikhin2021gshard}, where expert modules are selectively activated within individual network layers to efficiently scale model capacity, and these approaches have been particularly successful in large language models. In contrast, studying model-level MoEs remains important because it enables specialization across complete models with distinct inductive biases or training domains and provides clearer interpretability and modularity that are difficult to achieve when expert behavior is distributed across internal layers.

\begin{figure}[t]
  \centering
  
  \begin{subfigure}[b]{0.49\columnwidth}
    \centering
    \includegraphics[width=\textwidth]{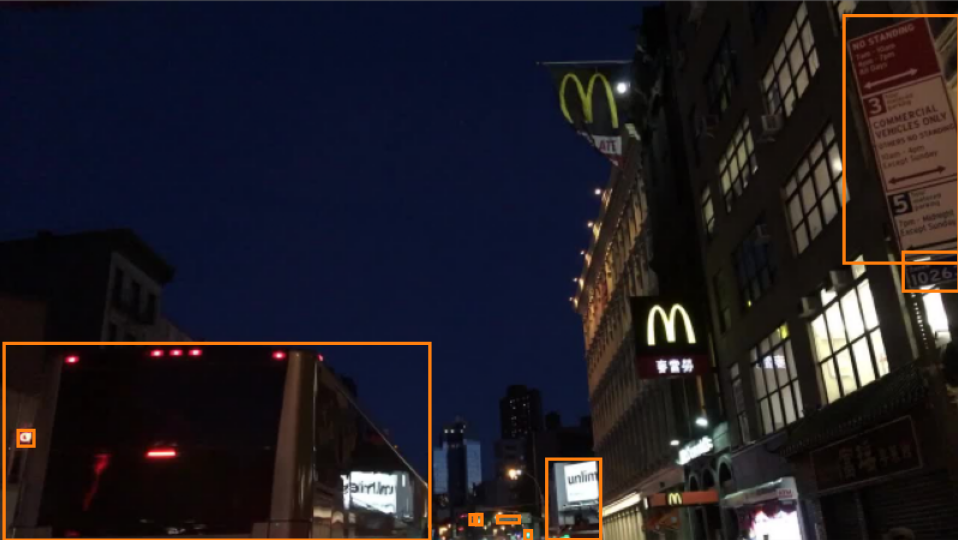}
    \caption*{Daytime expert}
  \end{subfigure}
    \begin{subfigure}[b]{0.49\columnwidth}
    \centering
    \includegraphics[width=\textwidth]{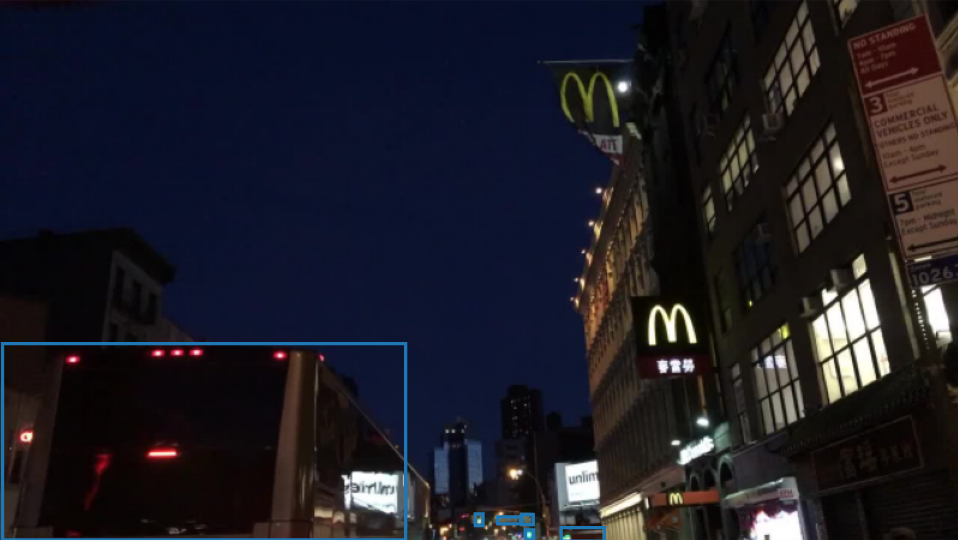}
    \caption*{Nighttime expert}
  \end{subfigure}

    \begin{subfigure}[b]{0.49\columnwidth}
    \centering
    \includegraphics[width=\textwidth]{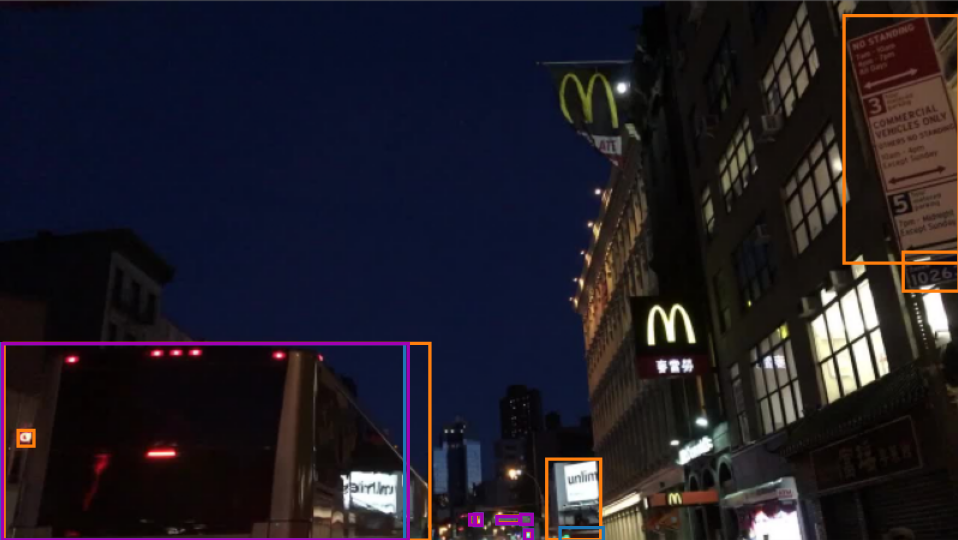}
    \caption*{Ensemble}
  \end{subfigure}
    \begin{subfigure}[b]{0.49\columnwidth}
    \centering
    \includegraphics[width=\textwidth]{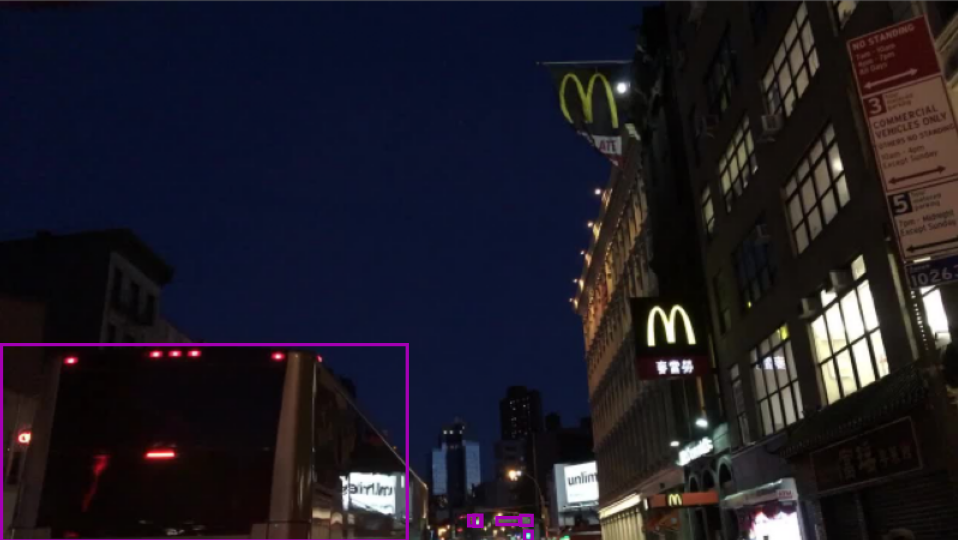}
    \caption*{MoE}
  \end{subfigure}

  \caption{Predictions of different models on an exemplary image from the BDD100K dataset.}
  \label{fig:example}
\end{figure}

Model-level MoE architectures have been extensively studied in image classification and semantic segmentation, where predictions are dense and structurally aligned~\cite{Valada2016a,pavlitskaya2020using,ahmed2016network,mees2016choosing}. In contrast, their application to object detection has received limited attention due to challenges in combining structured and variable-length predictions, such as bounding boxes (BBoxes) and confidence scores. These challenges make it nontrivial to apply expert weighting prior to detection postprocessing while preserving the benefits of specialization across experts.

In this work, we investigate model-level MoEs for object detection and analyze their effectiveness as an alternative to conventional ensembling. We propose an MoE architecture that combines  YOLO~\cite{redmon2016you}-based detectors trained on semantically disjoint data subsets, using a learned gating network that dynamically weights expert contributions at inference time. To address detection-fusion challenges, expert weighting is applied prior to postprocessing, and several strategies for merging BBox predictions are evaluated.

Experiments are conducted on the BDD100K dataset~\cite{yu2020bdd100k} using a split based on daytime and nighttime conditions to induce expert specialization. Results show that the proposed MoE consistently outperforms standard ensemble methods. 

Following the distinction between interpretability and explainability~\cite{doshi2017towards,rudin2019stop}, we focus on inherent interpretability, where the decision process is directly understandable from the model structure, and the proposed MoE provides it by explicitly exposing expert contributions, in contrast to ensembles that rely on opaque post hoc fusion. In addition to quantitative gains, our analysis of expert routing and prediction disagreement provides insight into domain-dependent expert behavior, demonstrating that model-level MoEs can improve both performance and interpretability in object detection.

\section{Related Work}

\subsection{Ensembles of CNNs for Object Detection}

Multiple strategies have been proposed for fusing predictions from several detectors. A common baseline aggregates BBox predictions from all models and applies class-level suppression to remove redundant detections. While this approach is efficient and architecture-agnostic, it may discard useful information when overlapping boxes are slightly misaligned. 

More advanced fusion methods explicitly combine spatially overlapping detections into a single prediction. All such approaches treat predictions from different detectors as if they originated from a single model and group BBoxes whose overlap exceeds a fixed threshold. Weighted Boxes Fusion (WBF) computes confidence weighted averages of BBox coordinates and adjusts the final score based on the number of contributing models~\cite{solovyev2021weighted}. Non Maximum Weighted (NMW) fusion applies weights derived from both confidence and spatial overlap with the most confident prediction in each group~\cite{zhou2017cad}. Soft NMS decays confidence scores in proportion to spatial overlap with higher scoring detections~\cite{bodla2017soft}. These methods allow consistent predictions across heterogeneous architectures to reinforce one another and often improve ensemble accuracy.

Following the approach of Xu et al.~\cite{xu2021forest}, ensemble predictions can be formed by aggregating final BBoxes from all experts and applying a suppression-based selection mechanism to produce a single prediction per object.

Casado et al. distinguish ensemble methods that operate within the detection algorithm from those that combine model outputs~\cite{casado2020ensemble}. Algorithm-level approaches include feature fusion prior to region proposal generation~\cite{li2017ensemble, peng2018megdet}, classification-stage ensembling~\cite{guo2015deep, cai2018cascade}, and methods that integrate multiple stages of the detection pipeline~\cite{lee2018ensemble}. Output-level ensembles combine predictions from different detectors, such as Fast R-CNN and Faster R-CNN~\cite{vo2018ensemble}, YOLO-based models~\cite{redmon2016you}, or RetinaNet and Mask R-CNN~\cite{solovyev2021weighted}.

Bahhar et al. present a wildfire and smoke detection framework that combines a voting-based classifier with a two-stage YOLO detector~\cite{wildfire}. Images are processed in parallel by both components, and localization is performed only when the classifier predicts the presence of fire or smoke using YOLOv5~\cite{ultralytics2021yolov5}. The method achieves strong precision and recall with fast inference suitable for edge deployment, but depends on high-quality imagery and may produce false positives in challenging scenarios~\cite{zhang2024yolo}.

Huayhongthong et al. propose an ensemble framework for incremental object detection to reduce training cost~\cite{huayhongthong2020incremental}. The system combines a pre-trained detector with a transferred model trained on new object classes using YOLO as the base architecture. An ensemble decision module selects the final prediction by comparing class probabilities from both models, enabling the addition of new classes without retraining the original detector~\cite{huayhongthong2020incremental}.

\subsection{Mixture of Experts}

Mixture of experts originates from the adaptive mixture of local experts framework by Jacobs et al.~\cite{jacobs1991adaptive}, where a learned gating function combines the outputs of multiple experts, each of which is a complete model that can specialize on different regions of the input space. In computer vision image classification, this model-level mixture idea has been explored as a learnable alternative to standard ensembles, and it has proven especially relevant for fine-grained recognition, where different experts can focus on complementary visual cues. A representative example is the mixture of granularity-specific experts' approach for fine-grained categorization, which explicitly trains experts at different granularity levels and fuses their predictions via learned weighting~\cite{zhang2019learning}.  Very recent work continues this direction for fine-grained visual classification with heterogeneous experts and collaborative gating mechanisms~\cite{yang2026fg}.

For dense prediction, our previous work~\cite{pavlitskaya2020using} applies model-level mixtures to semantic segmentation by training separate segmentation networks as experts and learning a gate to fuse their outputs, emphasizing interpretability and insight into expert behavior across scenes. A closely related and influential vision line is the work by Valada et al.~\cite{Valada2016a}, which uses a mixture of deep experts for robust semantic segmentation and also exemplifies a common application setting for model-level mixtures in vision: multisensor fusion, where experts operate on different modalities and the mixture learns how to combine them depending on the environment. 

This multisensor pattern also appears in object detection, for example, in adaptive multimodal fusion approaches that combine modality-specific expert detectors, such as RGB, depth, and optical flow, to improve robustness under changing conditions~\cite{mees2016choosing}.  MoE dynamically adapts to changing environment settings, assigning higher weights to the most confident expert. 

For standard single-modality object detection, MoCaE highlights that combining multiple complete detectors can be limited by expert miscalibration and proposes calibrated expert fusion to better exploit complementary strengths across detection benchmarks~\cite{oksuz2023mocae}. While MoCaE focuses on calibrating heterogeneous detectors to prevent confidence-dominated fusion, our work studies model-level mixtures with learned gating, highlighting domain-aware routing and interpretability as key challenges and opportunities for MoEs in object detection.

\newpage
\section{Approach}

Building upon our previous works~\cite{pavlitskaya2020using,pavlitskaya2022evaluating}, we propose an MoE architecture comprising several pre-trained CNNs for object detection, each fine-tuned on a subdomain of the input distribution. Their predictions are then combined dynamically during inference.

\subsection{MoE Architecture}

\begin{figure}[h]
    \centering
    \scalebox{0.48}{
        \def\svgwidth{2\columnwidth}
        \input{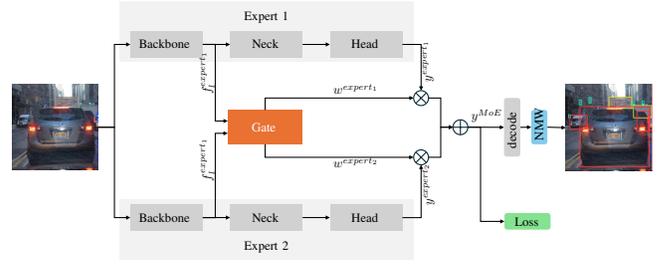}
    }
    \caption{Proposed MoE architecture consisting of two expert models.}
    \label{fig:moe_architecture}
\end{figure}

Formally, an MoE model consists of $n$ individual object detection experts
$\{F^{\text{expert}_i}\}_{i=1\ldots n}$ and a gating network $G$
(see Figure~\ref{fig:moe_architecture}). From each expert, a feature extraction
layer $\ell$ is selected, and the corresponding feature maps
$f_{\ell}^{\text{expert}_i}$ are concatenated and provided as input to the gate.
Given an input image $x \in \mathcal{I}^{H \times W \times C}$, where $H$ and $W$
denote the spatial dimensions, $C = 3$ the number of channels, and
$\mathcal{I} = [0,1]$, the gating network predicts a set of expert-specific
weights
\begin{equation}
\left(
w^{\text{expert}_1}, \ldots, w^{\text{expert}_n}
\right)
=
G\left(
f_{\ell}^{\text{expert}_i}
\big|_{i=1,\ldots,n}
\right).
\end{equation}

In classical MoE models, the gating network replaces the fixed averaging of expert predictions used in standard ensembles by predicting dynamic, input-dependent weights that control each expert’s contribution. Expert outputs are then multiplied by these weights rather than averaged as in an ensemble~\cite{jacobs1991adaptive,Valada2016a,pavlitskaya2020using}. However, object detectors produce structured
predictions that jointly encode BBox regression, objectness, and
class probabilities. Consequently, ensembles of object detectors do not perform explicit averaging. Instead, they concatenate predictions and rely on post-processing, such as NMS. To transfer the MoE principle to object detection, we therefore apply the gate to weight all expert predictions, including BBox coordinates and class scores, before decoding the predictions and applying a final fusion step such as NMS suppression. This design choice is necessary because weighting only BBox coordinates or only class scores would break the tight coupling between localization and classification in object detectors, leading to inconsistent detections and unreliable postprocessing.

Let $y^{\text{expert}_i}$ denote the raw output of expert
$i$, comprising all predicted BBox coordinates and associated scores.
Given the expert weights $w^{\text{expert}_i}$ predicted by the gate, the
weighted detector outputs are computed as
\begin{equation}
\tilde{y}^{\text{expert}_i}
=
w^{\text{expert}_i} \cdot y^{\text{expert}_i},
\quad i = 1, \ldots, n .
\end{equation}

The weighted predictions are then decoded using a shared anchor configuration
to recover BBox coordinates in image space. Finally, the decoded
predictions from all experts are merged using a post-processing step, such as
NMS, to obtain the final set of detections.

\subsection{MoE Training}

For MoE training, the expert weights are frozen, and only the gate is trained with the same loss as used for single models.

\textbf{Balancing loss}: To prevent expert collapse and encourage balanced expert utilization, several existing auxiliary balancing losses are used. For this, expert importance is defined as the cumulative contribution of each expert across a training batch as measured by the gating mechanism~\cite{shazeer2017outrageously}. For a given expert, this quantity is obtained by summing the gating weights assigned to that expert over all samples in the batch. Experts that receive consistently large gating weights are therefore considered more important under this definition.

The importance loss~\cite{shazeer2017outrageously} is designed to mitigate imbalanced expert utilization by penalizing large disparities in expert importance across the batch. It measures how unevenly the total gating mass is distributed among experts and applies a stronger penalty as the variation between experts increases. By minimizing this objective, the model is encouraged to use experts more evenly.

A probabilistic alternative is the KL divergence loss~\cite{pavlitska2023sparsely}, which treats expert importance as a discrete probability distribution over experts. The importance values are first normalized such that they sum to one across all experts. The loss then measures the divergence between this normalized distribution and a uniform distribution. Minimizing this divergence ensures that each expert receives an equal share of the total routing probability at the batch level.

Entropy-based loss~\cite{pavlitska2025robust} follows a similar motivation but directly maximizes the entropy of the normalized importance distribution. Higher entropy corresponds to a more uniform allocation of routing probability across experts. 

This objective promotes balanced expert usage when aggregated over the batch but does not explicitly constrain the sharpness of routing decisions for individual samples. To address this limitation, a sample-wise entropy regularization term is applied before batch aggregation. Instead of operating on batch-level importance, this loss penalizes low-entropy gating distributions at the sample level. As a result, the gating network is discouraged from making overly confident routing decisions for individual inputs. The final training objective combines the task loss with this entropy regularization term, weighted by a coefficient $\lambda$.

\textbf{Domain-aware gate training:} The standard formulation uses the object detection loss, so the gate is not forced to route inputs to the experts, which correspond to the domain to which the input belongs. Instead, the MoE training aims to minimize object detection error. Domain-aware gate training uses cross-entropy loss on the input-domain labels, thereby enforcing routing inputs according to their domain.

\section{Experimental Setup}

We use an MoE containing two experts, trained on semantically disjoint subsets of data, inspired by our previous works~\cite{pavlitskaya2020using,pavlitskaya2022evaluating}. We used YOLO~\cite{redmon2016you}-based detectors as expert models because their one-stage design, anchor-based prediction mechanism, and efficient inference pipeline make them a widely adopted and representative class of modern object detectors for evaluating model-level MoEs.

\subsection{Dataset Split}
\begin{figure}[h]
    \centering
    \includegraphics[width=\linewidth]{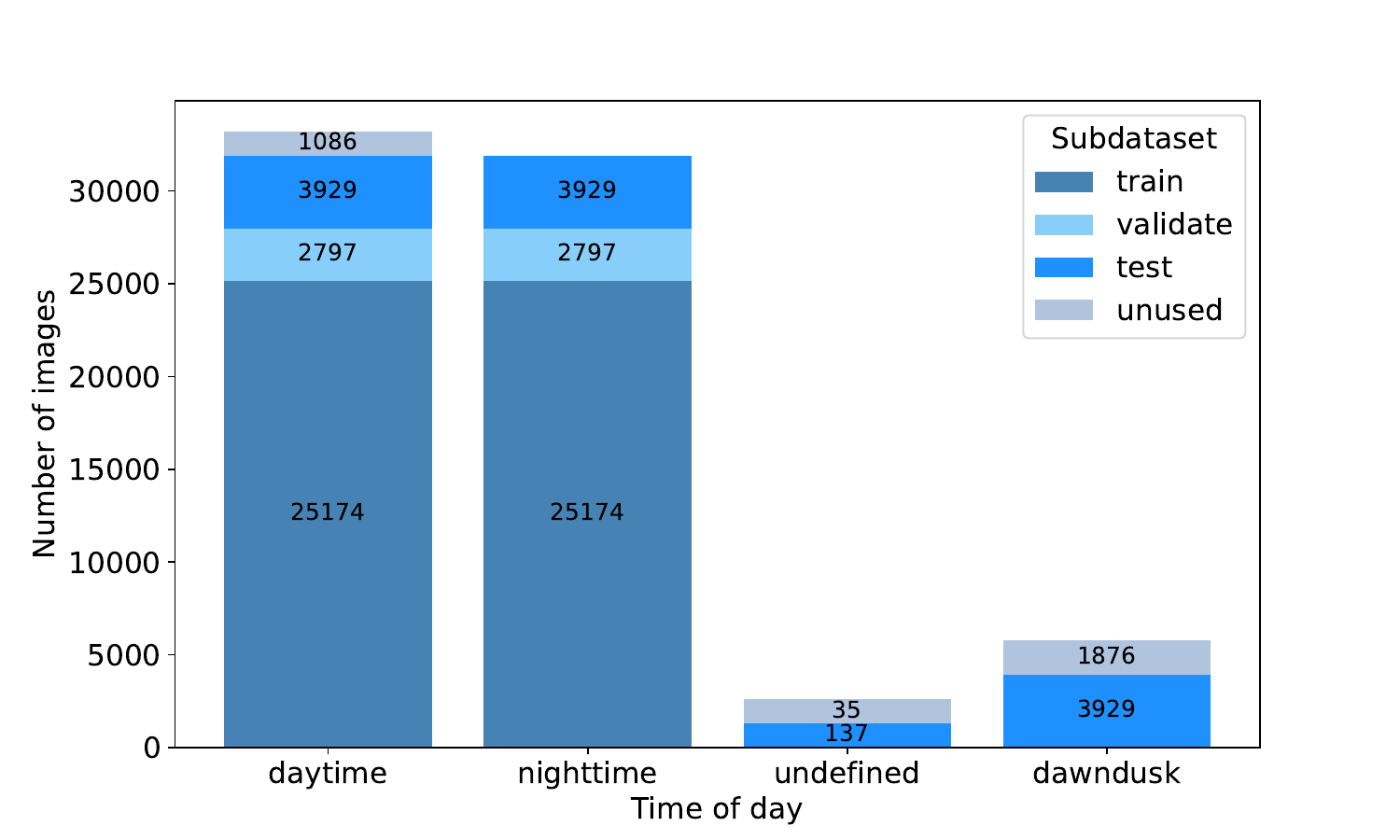}
    \caption{Distribution of BDD100K data according to the metadata parameter \textit{timeofday}.}
    \label{fig:timeofday-distr}
\end{figure}

\begin{figure}[h]
  \centering
  
  \begin{subfigure}[b]{0.49\columnwidth}
    \centering
    \includegraphics[width=\textwidth]{img/exemplary_images/day.jpg}
    \caption*{Daytime subset}
  \end{subfigure}
    \begin{subfigure}[b]{0.49\columnwidth}
    \centering
    \includegraphics[width=\textwidth]{img/exemplary_images/night3.jpg}
    \caption*{Nighttime subset}
  \end{subfigure}

    \begin{subfigure}[b]{0.49\columnwidth}
    \centering
    \includegraphics[width=\textwidth]{img/exemplary_images/dawn.jpg}
    \caption*{Dawn/dusk subset (only test)}
  \end{subfigure}
    \begin{subfigure}[b]{0.49\columnwidth}
    \centering
    \includegraphics[width=\textwidth]{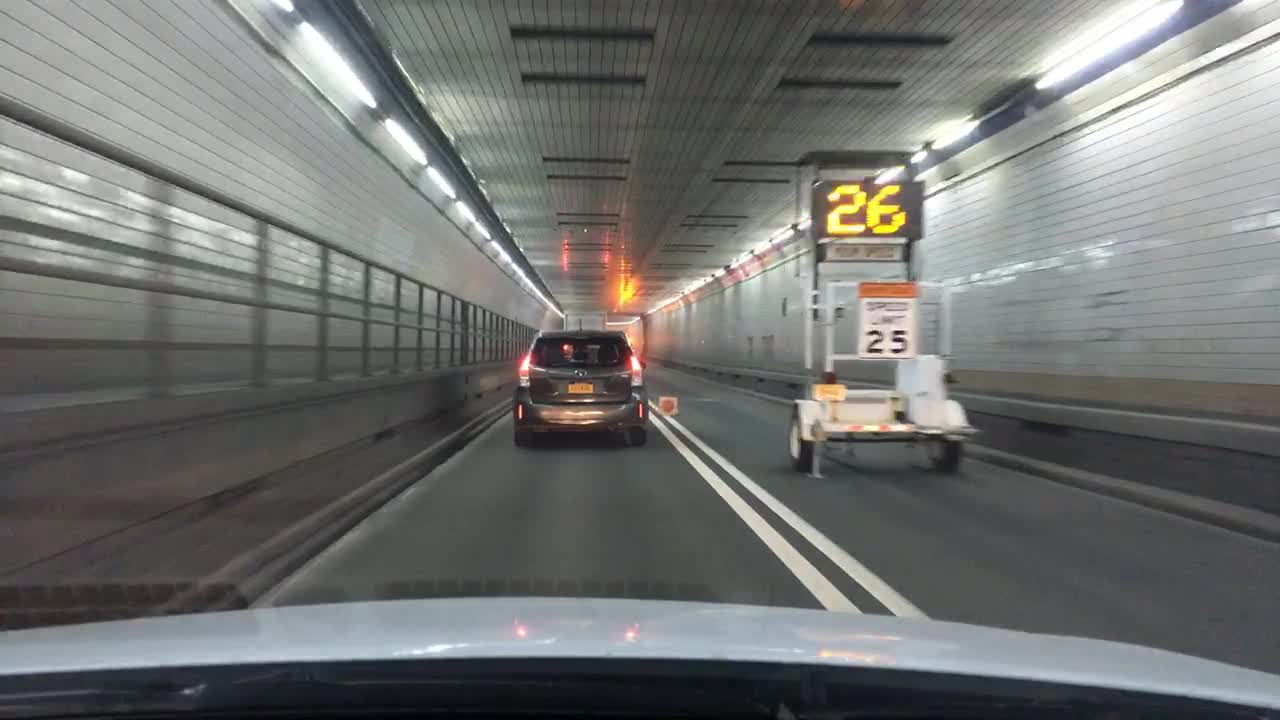}
    \caption*{Undefined subset (only test)}
  \end{subfigure}

  \caption{Exemplary images with different values of the \textit{timeofday} metadata parameter in the BDD100K dataset}
  \label{fig:images}
\end{figure}

We use the Berkeley DeepDrive 100K (BDD100K) dataset~\cite{yu2020bdd100k}, which we split according to the \textit{time of day} metadata parameter (see Figures~\ref{fig:timeofday-distr} and~\ref{fig:images}). We use \textit{nighttime} data to train the \textit{nighttime} expert model, and \textit{daytime} data to train the \textit{daytime} expert model. The BDD100K dataset contains approximately 35K daytime and 27K nighttime images. 27,971 images are randomly selected to train each expert, ensuring balanced training and eliminating bias. Given the absence of labeled test data in the BDD100K dataset, the training images are split at 90:10, resulting in 25,174 training and 2,798 validation images. 

The original validation data is used as test data, with 3,929 test images per subset. In addition to \textit{daytime} and \textit{nighttime} data, we report results on 3,939 \textit{dawn/dusk} images, and 1,876 \textit{undefined} images.

\subsection{Training Settings}
We used open-source PyTorch implementations of YOLOv7~\cite{wang2023yolov7}\footnote{https://github.com/WongKinYiu/yolov7} and YOLOv9~\cite{wang2024yolov9}\footnote{https://github.com/WongKinYiu/yolov9}. 
For each version, one large and one medium-sized variant were chosen. Models were trained
for 200 epochs and evaluated with a confidence threshold of 0.001 and an IoU threshold of 0.6 at a NVIDIA RTX 2080 Ti GPU with 11GB VRAM. 

Each expert model was trained on the corresponding subset. The baseline and the MoE were trained on the combined \textit{daytime-nighttime} train data. During MoE training, expert weights were frozen. We report mAP50 on the test data for all models.

The anchor configurations for all YOLOv7 models were optimized on the combined training data prior to training. As these anchors are also used for scaling during postprocessing, we use identical anchors for all models, optimized on the combined domains. For YOLOv9 inference, the primary detection head was used.

For the entropy balancing loss in an MoE, the regulation weight $\lambda$ was set to 0.1 for YOLOv7 and 0.3 for YOLOv9.

\section{Evaluation}

\subsection{Detection Performance}

\textbf{Performance of single models:} We have compared the detection performance of single models, ensembles, and an MoE (see Table~\ref{tab:performance}). Expert models perform the best on the corresponding subset they were trained on. In most cases, baselines outperform the expert models.

\begin{table}[t]
\caption{Model comparison over data subsets.}
\centering
  \resizebox{1.0\linewidth}{!}{
    \begin{tabular}{|r| c c  c  c  c c | c| c|}
    \hline
    \textbf{Model} &  \multicolumn{6}{c|}{\textbf{mAP $\uparrow$ (\%) on data subsets}} & \textbf{Num.}& \textbf{Inf.}\\

    & \textbf{Day} & \textbf{Night} & \textbf{Dawn/} & \textbf{Undef.} & \textbf{Day+} & \textbf{All} & \textbf{params,} & \textbf{time,}\\
    &  &  & \textbf{Dusk} &  & \textbf{Night} & & \textbf{M} & \textbf{ms}\\\hline

    \rowcolor{LightBlue}
    \multicolumn{9}{|c|}{\textbf{YOLOv7-tiny, trained from scratch}}\\ 
    Baseline & 46.39 & \textbf{45.80} & \textbf{45.87} & \textbf{69.30} & \textbf{46.30} & \textbf{46.16} & 6.2 & 82 \\
    Daytime expert & \textbf{46.53} & 34.92 & 45.00 & 56.33 & 42.78 & 43.50 & 6.2 & 82 \\
    Nighttime expert & 32.62 & 43.20 & 34.62 & 59.60 & 36.55 & 35.86 & 6.2 & 82 \\ \hline
    Ensemble & 45.97 & 42.67 & 45.18 & 61.57 & 44.90 & 45.10 & 12.4 & 186 \\
    MoE & 46.54 & 43.82 & 45.52 & 60.36 & 45.97 & 45.79 & 12.4 & 82 \\
    \hline

    \rowcolor{LightBlue}
    \multicolumn{9}{|c|}{\textbf{YOLOv7, pre-trained on COCO}}\\ 
    Baseline & 59.68 & \textbf{56.51} & \textbf{59.31} & 78.65 & \textbf{58.42} & \textbf{58.70} & 63.9 & 73 \\
    Daytime expert & \textbf{60.07} & 49.46 & 58.00 & \textbf{76.78} & 56.39 & 56.95 & 63.9 & 73 \\
    Nighttime expert & 49.16 & 54.42 & 50.96 & 72.37 & 50.73 & 50.80 & 63.9 & 73 \\ \hline
    Ensemble NMW & 59.65 & 54.38 & 58.34 & 78.05 & 57.52 & 57.90 & 127.8 & 157 \\
    MoE & \textbf{60.07} & 54.75 & 58.40 & 77.24 & 58.11 & 58.21 & 127.8 & 80 \\
    \hline

    \rowcolor{LightBlue}
    \multicolumn{9}{|c|}{\textbf{YOLOv7x, pre-trained on COCO}}\\

    Baseline & \textbf{60.82} & \textbf{57.08} & \textbf{60.08} & \textbf{78.78} & \textbf{59.45} & \textbf{59.66} & 71.3 & 82 \\ 
    Daytime expert & 60.47 & 50.40 & 58.94 & 76.26 & 57.07 & 57.66 & 71.3 & 82 \\ 
    Nighttime expert & 51.26 & 54.79 & 52.30 & 67.81 & 52.27 & 52.21 & 71.3 & 82 \\ \hline
    Ensemble & 60.54 & 54.26 & 59.48 & 77.80 &  58.36 & 58.69 & 142.6 & 170 \\
    MoE & \textbf{60.82} & 55.04 & 59.39 & 74.38 & 58.75 & 58.95 & 142.6 & 97  \\
    \hline

    \rowcolor{LightBlue}
    \multicolumn{9}{|c|}{\textbf{YOLOv7x, trained from scratch }}\\ 
    Baseline & \textbf{59.79} & \textbf{56.06} & \textbf{59.18} & \textbf{76.84} & \textbf{58.43} & \textbf{58.64} & 71.3 & 83 \\ 
    Daytime expert & 59.49 & 49.28 & 57.73 & 74.29 & 56.11 & 56.64 & 71.3 & 83 \\ 
    Nighttime expert & 48.34 & 53.34 & 49.95 & 68.33 & 49.83 & 49.83 & 71.3 & 83 \\  \hline
    Ensemble & 59.09 & 53.45 & 58.32 & 76.39 & 57.12 & 57.55 & 142.6 & 178 \\
    MoE & 59.52 & 53.86 & 57.96 & 76.01 & 57.60 & 57.68 & 142.6 & 96 \\
    \hline

    \rowcolor{LightBlue}
    \multicolumn{9}{|c|}{\textbf{YOLOv9-S, pre-trained on COCO}}\\ 
    Baseline & \textbf{51.43} & \textbf{50.26} & \textbf{51.33} & \textbf{66.83} & \textbf{51.04} & \textbf{51.08} & 7.1 & 54 \\
    Daytime expert & 51.15 & 40.00 & 49.93 & 66.03 & 47.53 & 48.27 & 7.1 & 54 \\
    Nighttime expert & 38.76 & 46.28 & 40.54 & 49.24 & 41.45 & 41.04 & 7.1 & 54 \\ \hline
    Ensemble & 50.14 & 45.91 & 49.79 & 68.27 & 48.82 & 49.09 & 14.2 & 73 \\
    MoE & 51.31 & 47.20 & 50.71 & 64.83 & 50.16 & 50.24 & 14.2 & 76 \\
    \hline

    \rowcolor{LightBlue}
    \multicolumn{9}{|c|}{\textbf{YOLOv9c, pre-trained on COCO}}\\ 
    Baseline & 56.83 & \textbf{54.38} & \textbf{56.69} & \textbf{73.52} & \textbf{55.91} & \textbf{56.15} & 25.3 & 63 \\ 
    Daytime expert & \textbf{56.95} & 49.70 & 55.53 & 69.37 & 54.41 & 54.53 & 25.3 & 63 \\ 
    Nighttime expert & 47.22 & 52.07 & 49.25 & 64.49 & 48.80 & 48.86 & 25.3 & 63 \\  \hline
    Ensemble & 56.50 & 52.66 & 56.08 & 74.44 & 55.07 & 55.29 & 50.6 & 98 \\
    MoE & 57.75 & 53.17 & 56.89 & 73.73 & 55.85 & 56.13 & 50.6 & 94 \\
    \hline

    \rowcolor{LightBlue}
    \multicolumn{9}{|c|}{\textbf{YOLOv9c, trained from scratch}}\\  
    Baseline & \textbf{56.68} & \textbf{53.49} & \textbf{55.97} & \textbf{71.23} & \textbf{55.65} & \textbf{55.72} & 25.3 & 63 \\
    Daytime expert & 56.03 & 45.74 & 54.40 & 69.04 & 52.71 & 53.21 & 25.3 & 63 \\ 
    Nighttime expert & 44.99 & 50.52 & 46.45 & 67.71 & 46.85 & 46.65 & 25.3 & 63 \\  \hline
    Ensemble & 55.61 & 50.95 & 54.89 & 73.73 & 54.12 & 54.34 & 50.6 & 98 \\ 
    MoE & 56.24 & 51.25 & 55.50 & 71.72 & 54.66 & 54.85 & 50.6 & 93 \\
    \hline
    
    \rowcolor{LightBlue}
    \multicolumn{9}{|c|}{\textbf{YOLOv9-E, pre-trained on COCO}}\\ 
    Baseline & 59.16 & \textbf{56.02} & \textbf{59.04} & 73.61 & \textbf{58.01} & \textbf{58.27} & 57.3 & 69 \\
    Daytime expert & \textbf{59.48} & 51.23 & 60.50 & \textbf{74.37} & 56.61 & 57.89 & 57.3 & 69 \\
    Nighttime expert & 50.28 & 53.35 & 51.59 & 68.48 & 51.11 & 51.18 & 57.3 & 69 \\ \hline
    Ensemble & 59.21 & 54.49 & 60.67 & 77.12 & 57.43 & 58.50 & 114.6 & 107 \\
    MoE & 59.93 & 55.36 & 61.41 & 75.24 & 58.12 & 59.19 & 114.6 & 107 \\
    \hline




    \end{tabular}
}
\label{tab:performance}
\end{table}

\textbf{Ensemble variants:} An ensemble of two experts was used as a comparison approach for the proposed MoE. For an ensemble, we have compared several methods for fusing the outputs of individual models (see Table~\ref{tab:ensembles}). For SoftNMS~\cite{bodla2017soft}, $\sigma=0.5$ and $\tau=0.3$ were used. NMW~\cite{zhou2017cad} showed the best mAP for all model variants. Across all methods, pre-weighting BBox confidence scores before applying them based on each model's relative mAP further increased mAP. Therefore, NMW with mAP pre-weighting was used for all ensemble models.

\begin{table}[h]
\caption{Output fusion approaches, evaluated for a YOLOv7x ensemble.}
\centering
    \begin{tabular}{|r| c | c|}
    \hline
    \textbf{Algorithm} &  \multicolumn{2}{c|}{\textbf{mAP $\uparrow$ (\%)}} \\
    & \textbf{Standard} & \textbf{With mAP re-weighting } \\ 
    \hline
    NMS & 58.29 & 58.53\\
    WBF~\cite{solovyev2021weighted} & 56.09 & 55.94\\
    NMW~\cite{zhou2017cad} & 58.45 &  \textbf{58.69}\\
    SoftNMS~\cite{bodla2017soft}  & 41.16 & 42.38\\
    \hline
    \end{tabular}
\label{tab:ensembles}
\end{table}

\textbf{MoE performance:} The results (see Table~\ref{tab:performance}) show that MoEs consistently outperform fixed ensembles across model variants, indicating that learned routing is more effective than static fusion of expert predictions. However, MoEs generally do not surpass the strongest baseline models trained on the full, joint dataset, which benefit from exposure to the complete data distribution. Table~\ref{tab:improvement} summarizes the relative mAP improvement compared to single experts and ensembles. The largest relative improvement could be reached on in-domain nighttime and daytime data.

\begin{table}[t]
\caption{Relative mAP improvement of YOLOv7x-based MoE compared to other models.}
\centering
  \resizebox{1.0\linewidth}{!}{
     \begin{tabular}{|r| c c  c  c  c | }

   \hline
    & \textbf{Day} & \textbf{Night} & \textbf{Dawn/Dusk} & \textbf{Undef.}  & \textbf{All}\\\hline

    Compared to experts  & \textbf{0.1565} & 0.1004 & 0.1281 & 0.0990 & 0.1279  \\
    Compared to ensemble  & 0.0111 & \textbf{0.0164} & 0.0106 & 0.0094 & 0.0127  \\
   \hline

    \end{tabular}
}
\label{tab:improvement}
\end{table}

\textbf{Performance over data subdomains:} The domain-specific experts achieve their strongest performance on the subsets they were trained on (daytime or nighttime) but degrade noticeably on dawn/dusk and undefined images, indicating limited generalization beyond their specialization. We consider the dawn/dusk and undefined subsets as out-of-distribution (OOD) data since they are not included in the expert training splits and exhibit systematically lower detection performance across all models, indicating domain characteristics that differ from the training distribution. The baseline models are more robust on these OOD subsets and generally outperform individual experts, reflecting their broader exposure during training. Standard ensembles improve robustness over single experts by aggregating predictions, but their gains on OOD data remain limited (see Table~\ref{tab:performance}). The MoE consistently matches or slightly exceeds ensemble performance on these subsets and, in several configurations, approaches the baseline performance. 

\begin{table}[t]
\caption{MoE ablation studies for YOLOv7x.}
\centering
  \resizebox{1.0\linewidth}{!}{
     \begin{tabular}{|r| c c  c  c  c  c | }
    \hline
    \textbf{Model} &  \multicolumn{6}{c|}{\textbf{mAP $\uparrow$ (\%) on data subsets}} \\

    & \textbf{Day} & \textbf{Night} & \textbf{Dawn/} & \textbf{Undef.} & \textbf{Day+} & \textbf{All}\\
    &  &  & \textbf{Dusk} &  & \textbf{Night} & \\\hline

    

    \rowcolor{LightBlue}
    \multicolumn{7}{|c|}{\textbf{No domain-aware routing, different balancing losses}}\\ 

    No balancing loss & 60.47 & 54.20 & 58.95 & \textbf{75.81} & 58.34 & 58.48  \\ 
    Importance loss & 60.42 & 54.68 & 58.95 & 70.26 & 58.42 & 58.51  \\
    KL loss & 60.47 & 54.20 & 58.95 & \textbf{75.81} & 58.34 & 58.48  \\
    Batch-wise entropy loss & 60.45 & 54.79 & 58.95 & 69.65 & 58.46 & 58.53 \\
    Sample-wise entropy loss & \textbf{60.82} & \textbf{55.04} & \textbf{59.39} & 74.38 & \textbf{58.75} & \textbf{58.95}  \\ \hline
    
    \rowcolor{LightBlue}
    \multicolumn{7}{|c|}{\textbf{Domain-aware gate training, different balancing losses}}\\ 
    No balancing loss & 60.45 & \textbf{54.95} & 59.31 & \textbf{75.79} & 58.50 & \textbf{58.71}  \\ 
    Importance loss & 60.32 & 54.90 & 59.34 & 70.02 & 58.44 & 58.68    \\ 
    KL loss loss & \textbf{60.49} & \textbf{54.95} & 59.29 & 70.63 & \textbf{58.52} & 58.70    \\ 
    Batch-wise entropy loss & 60.47 & 54.86 & 59.24 & 73.90 & 58.49 & 58.67    \\ 
    Sample-wise entropy loss & 60.46 & 54.90 & \textbf{59.33} & 75.75 & 58.49 & \textbf{58.71}    \\ \hline

    \rowcolor{LightBlue}
    \multicolumn{7}{|c|}{\textbf{Different gate architectures}}\\ 
   
    1 FC layer & \textbf{60.83} & 54.93 & 59.31 & 70.01 & 58.71 & 58.79 \\
    2 FC layers & 60.48 & 54.97 & \textbf{59.40} & \textbf{75.75} & 58.53 &  58.76 \\
    Conv and 2 FC layers  & 60.82 & \textbf{55.04} & 59.39 & 74.38 & \textbf{58.75} & \textbf{58.95}    \\
    2 Conv and 2 FC layers & 60.80 & 55.14 & 59.37 & 77.14 & 58.74 & 58.88 \\
    
    \hline

    \rowcolor{LightBlue}
    \multicolumn{7}{|c|}{\textbf{Different number of gates}}\\ 
    Single & \textbf{60.82} & \textbf{55.04} & \textbf{59.39} & 74.38 & \textbf{58.75} & \textbf{58.95}  \\
    Spatial  & 60.53 & 55.13 & 59.32 & 76.94 & 58.67 & 58.87 \\
    Class-wise & 60.37 & 53.29 & 59.14 & 76.28 & 58.01 & 58.38  \\

    \hline
    
    \rowcolor{LightBlue}
    \multicolumn{7}{|c|}{\textbf{Different feature extraction layers}} \\
    Early layer (L43) & \textbf{60.83} & \textbf{55.11} & 59.35 & \textbf{76.46} & \textbf{58.74} & \textbf{58.86} \\
    Middle layer (L58) & 60.82 & 55.04 & 59.39 & 74.38 & 58.75 & 58.95  \\ 
    Late layer (L117)& 60.55 & 54.90 & \textbf{59.51} & 70.30 & 58.54 & 58.78 \\
    \hline
    
    \rowcolor{LightBlue}
    \multicolumn{7}{|c|}{\textbf{Fixed weights instead of routing}} \\
    Equal-weighted     & 57.34 & 54.14 & 57.75 & 73.31 & 56.20 & 56.69 \\
    mAP-weighted       & 57.74 & 53.90 & 58.17 & 75.00 & 56.41 & 56.98 \\
    Routing & \textbf{60.82} & \textbf{55.04} & \textbf{59.39} & \textbf{74.38} & \textbf{58.75} & \textbf{58.95}  \\ 
    \hline


    \end{tabular}
}
\label{tab:ablations}
\end{table}

\subsection{Ablation studies}

\textbf{Training from scratch:} Model pre-trained on COCO~\cite{lin2014microsoft} showed higher mAP than models trained from scratch on the corresponding BDD100K subset (see Table~\ref{tab:performance}). 

\textbf{Balancing loss:} We performed a grid search to optimize the scale of each loss and a decay factor relative to the training epoch. Among the evaluated losses, sample-wise entropy achieved the highest overall mAP. Functions were evaluated to prevent the MoE model from collapsing to a single expert, an effect observed in the ablation studies (see Table~\ref{tab:ablations}).


\textbf{Domain-aware training:} Using cross-entropy loss on the input-domain labels has performed worse than unsupervised training over all losses, but still better than a single expert.  A possible reason is that enforcing domain-based routing restricts the gate to coarse metadata labels and prevents it from exploiting finer-grained visual cues that are more informative for optimizing detection performance.

\textbf{Feature extraction layer:} Our gate uses features from the last layer of the backbone (the feature pyramid) as input from experts. It is referred to as a middle layer (see Table~\ref{tab:ablations}). In YOLO models, it corresponds to the highest level of abstraction and the lowest feature size. This makes it well-suited for a global, image-level gating decision. We also evaluated two alternative locations in YOLOv7: an earlier backbone layer (L58), providing features closer to the image, and the final neck layer (L117). The neck refines backbone features by upsampling and downsampling across pyramid levels while using skip connections to combine multi-scale information. Its final layer also has high abstraction and low feature count. Among all tested configurations, using the last backbone layer achieved the highest mAP.

\textbf{Gate architecture:} We evaluated four gate architectures with increasing complexity:

\begin{itemize}
    \item One FC layer: AvgPool $\rightarrow$ FC(2) $\rightarrow$ Softmax 
    \item Two FC layers: AvgPool $\rightarrow$ FC(512) $\rightarrow$ FC(2) $\rightarrow$ Softmax 
    \item One conv and two FC layers: Conv(1$\times$1) $\rightarrow$ AvgPool $\rightarrow$ FC(512) $\rightarrow$ FC(2) $\rightarrow$ Softmax 
    \item Two conv and two FC layers: Conv(1$\times$1) $\rightarrow$ BN $\rightarrow$ Conv(3$\times$3) $\rightarrow$ BN $\rightarrow$ AvgPool $\rightarrow$ FC(512) $\rightarrow$ FC(2) $\rightarrow$ Softmax 
\end{itemize}

The variant with one convolutional and two fully connected layers performed best (see Table~\ref{tab:ablations}). This architecture evidently provides sufficient representational capacity while remaining simple enough to avoid overfitting.


\textbf{Number of gates:} In addition to a single image-level gate, we also evaluated a classwise gate that consists of one simple gate per class and a shared gate for coordinates and objectness, allowing class scores to be merged independently of box coordinates. In addition, we evaluated a spatial gate with one gate per spatial feature location (e.g., 20×20 in YOLOv7). These gate weights are bilinearly interpolated for each output layer and merge both coordinates and class scores for the corresponding image region. The single gate achieved the highest mAP and was therefore chosen for other experiments.

\textbf{Fixed weights:}
Omitting the gates and using fixed expert weights instead of the predicted ones resulted in lower mAP (see Table~\ref{tab:ablations}), thus stressing the benefit of using MoEs.

\subsection{Impact on Interpretability}

In this work, interpretability is defined as the ability to attribute detection outcomes to individual experts and to analyze how expert contributions vary across input conditions. For object detection, this includes understanding which expert dominates a prediction, how expert confidence changes across domains, and how disagreement between experts relates to final detection decisions. An interpretable model, therefore, exposes both global patterns of expert specialization across datasets and local decision behavior at the level of individual images and detected objects.

\begin{figure}[h]
  \centering
  
  \begin{subfigure}[b]{0.49\columnwidth}
    \centering
    \includegraphics[width=\textwidth]{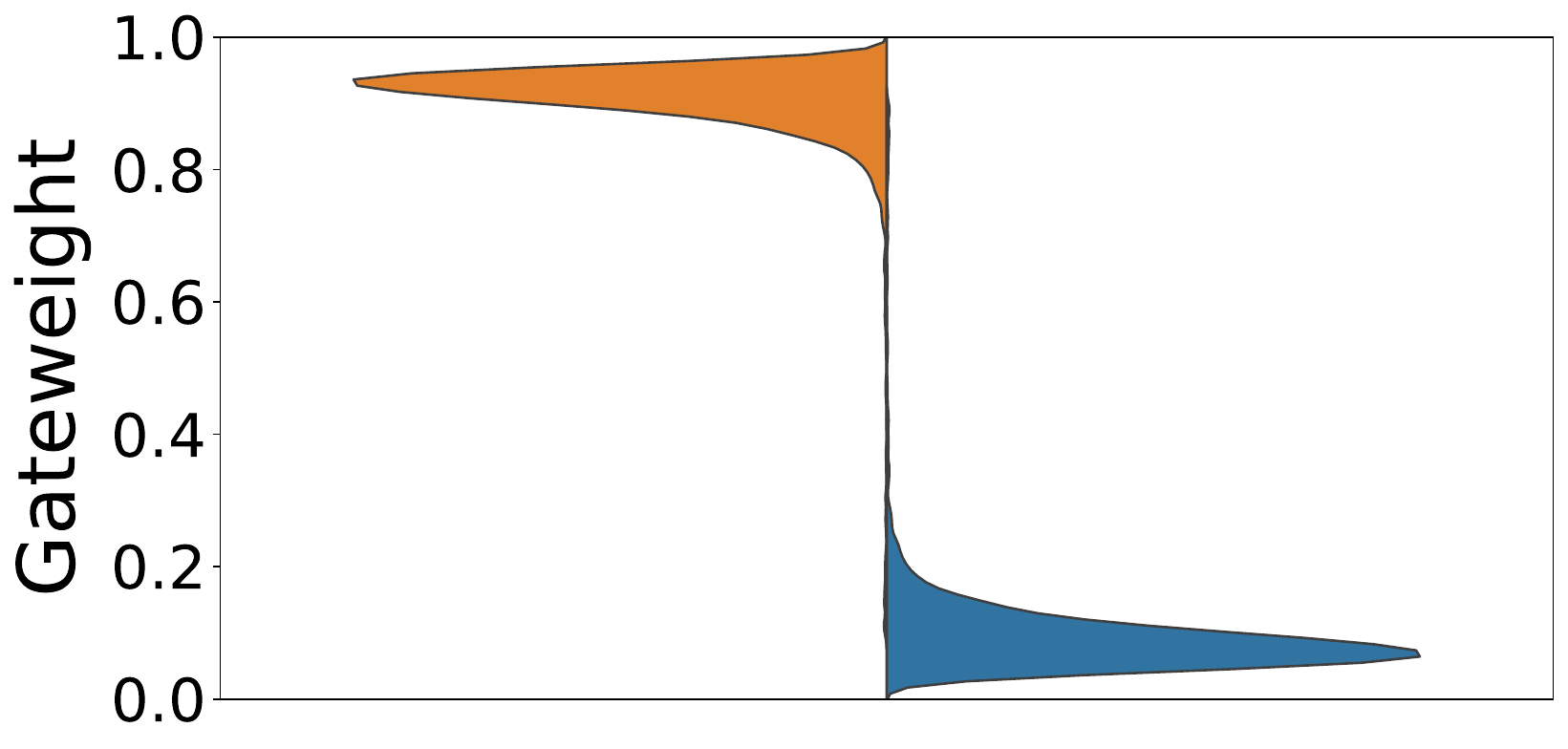}
    \caption*{Daytime subset}
  \end{subfigure}
    \begin{subfigure}[b]{0.49\columnwidth}
    \centering
    \includegraphics[width=\textwidth]{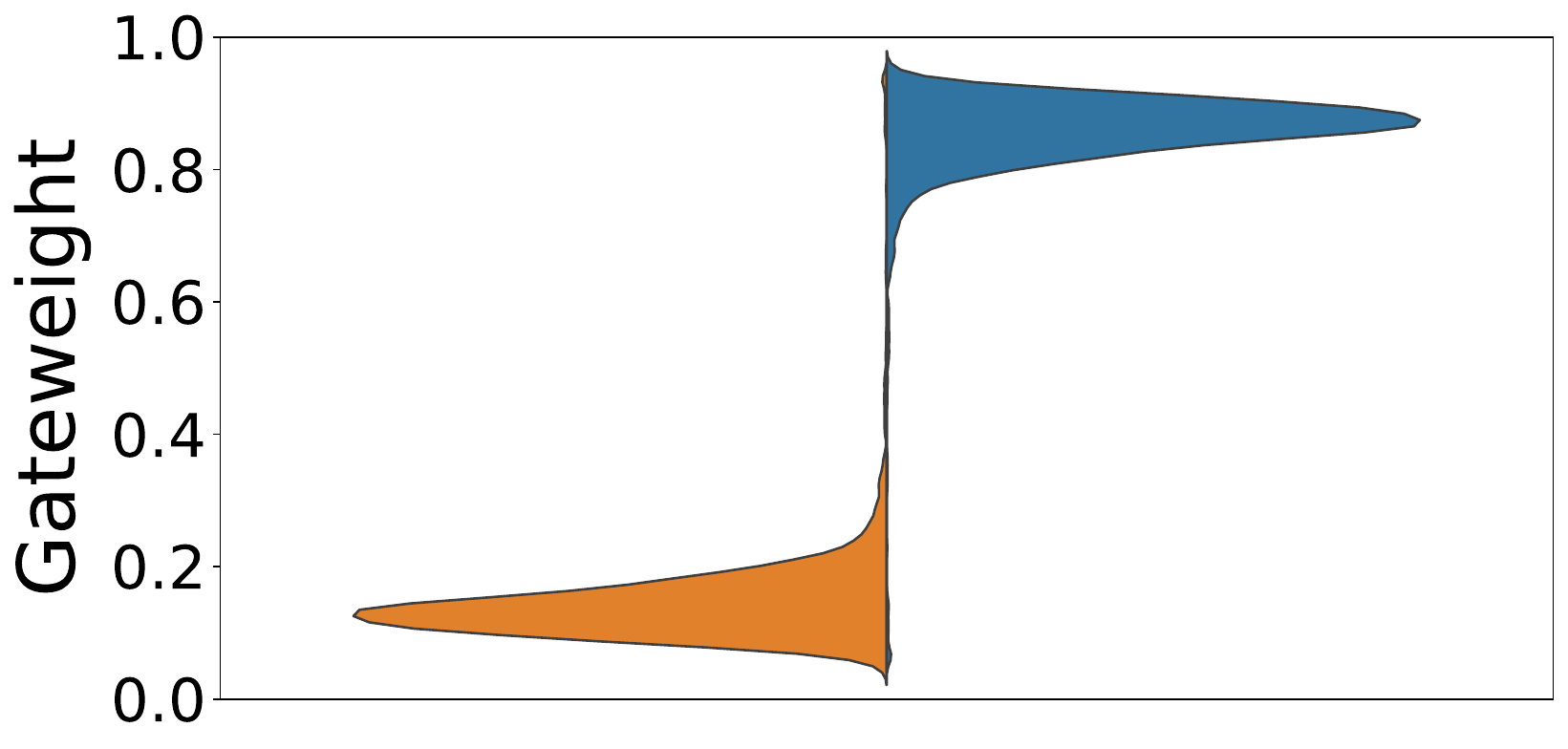}
    \caption*{Nighttime subset}
  \end{subfigure}
  
  \begin{subfigure}[b]{0.49\columnwidth}
    \centering
    \includegraphics[width=\textwidth]{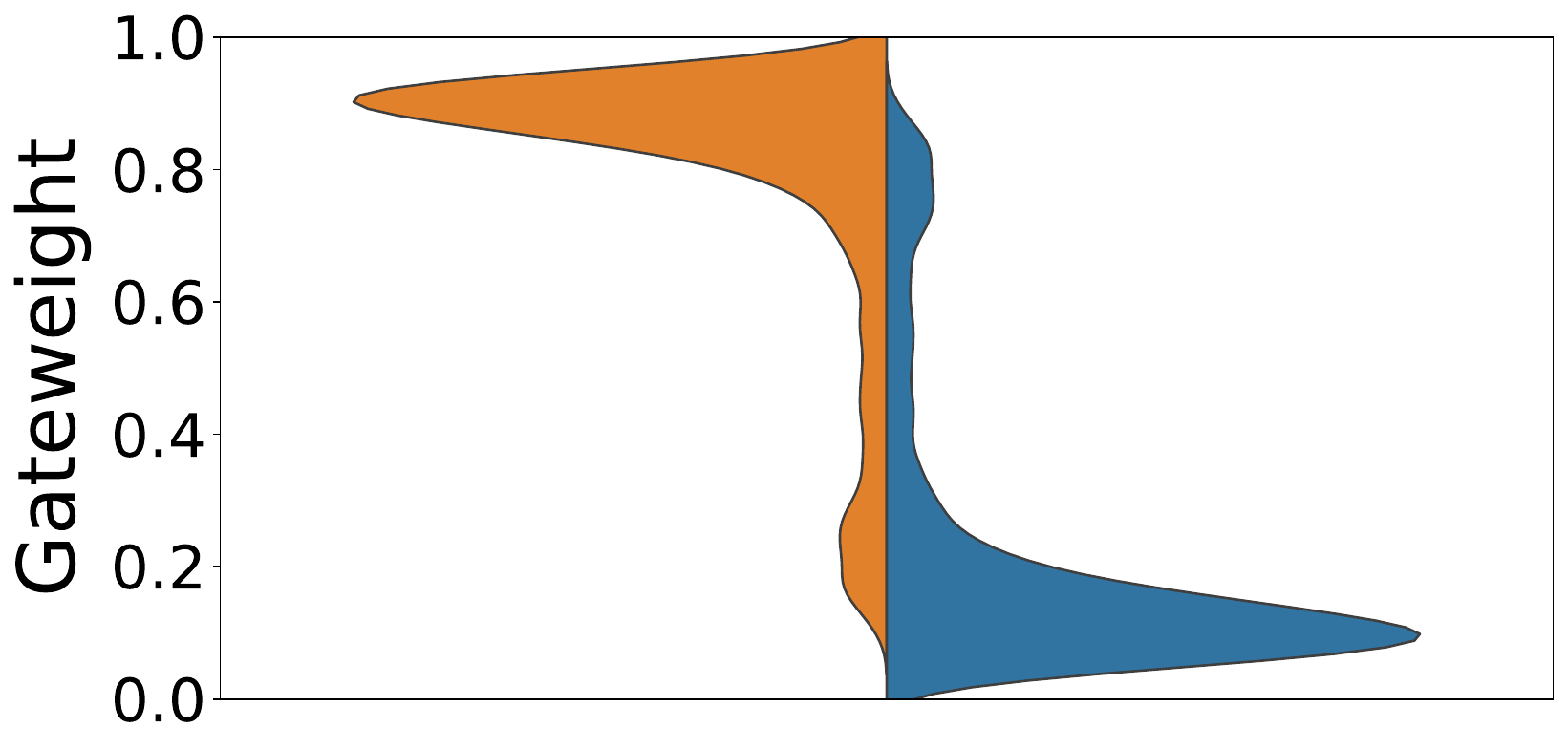}
    \caption*{Dawn/dusk subset}
  \end{subfigure}
    \begin{subfigure}[b]{0.49\columnwidth}
    \centering
    \includegraphics[width=\textwidth]{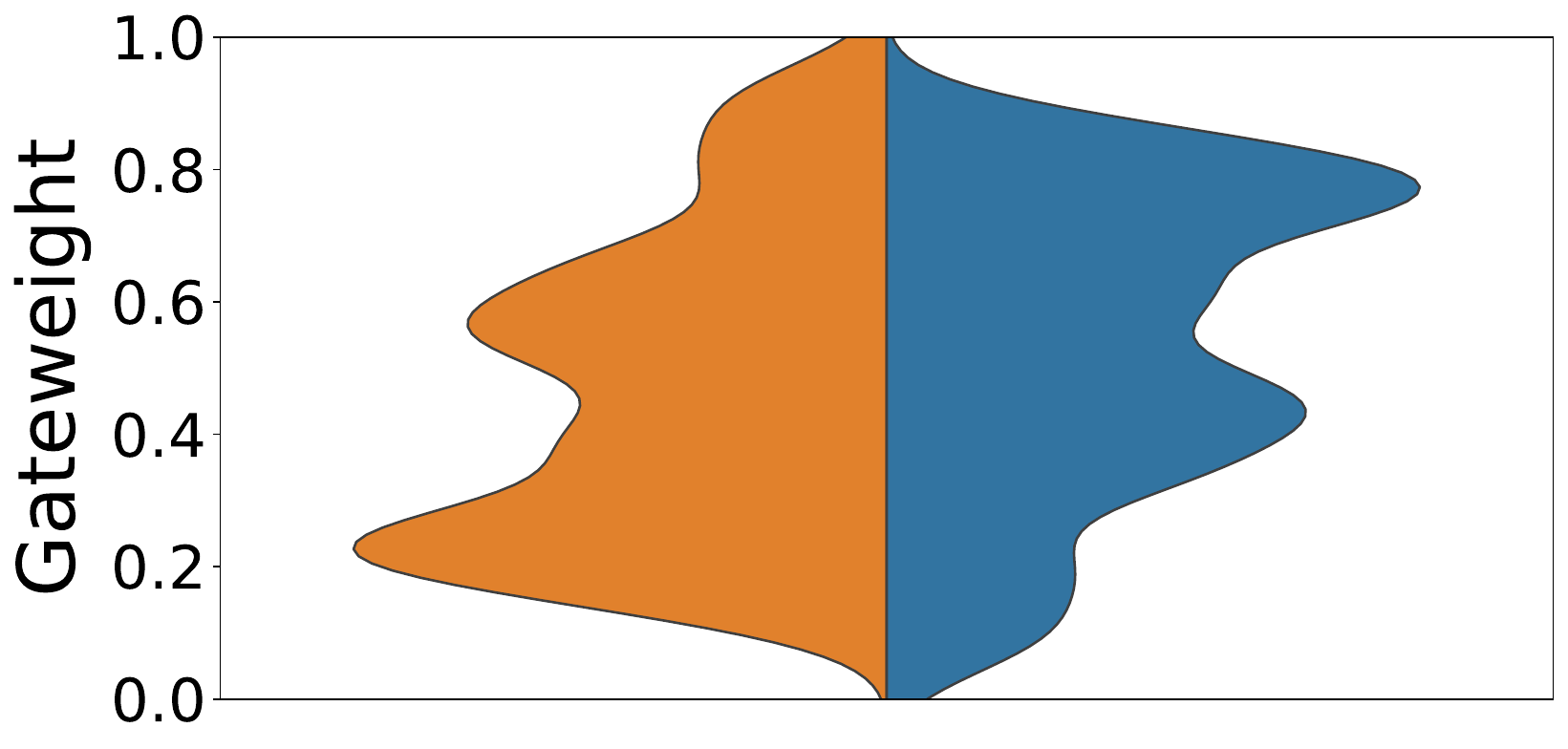}
    \caption*{Undefined subset}
  \end{subfigure}
  
\begin{subfigure}[b]{0.49\columnwidth}
    \centering
    \includegraphics[width=\textwidth]{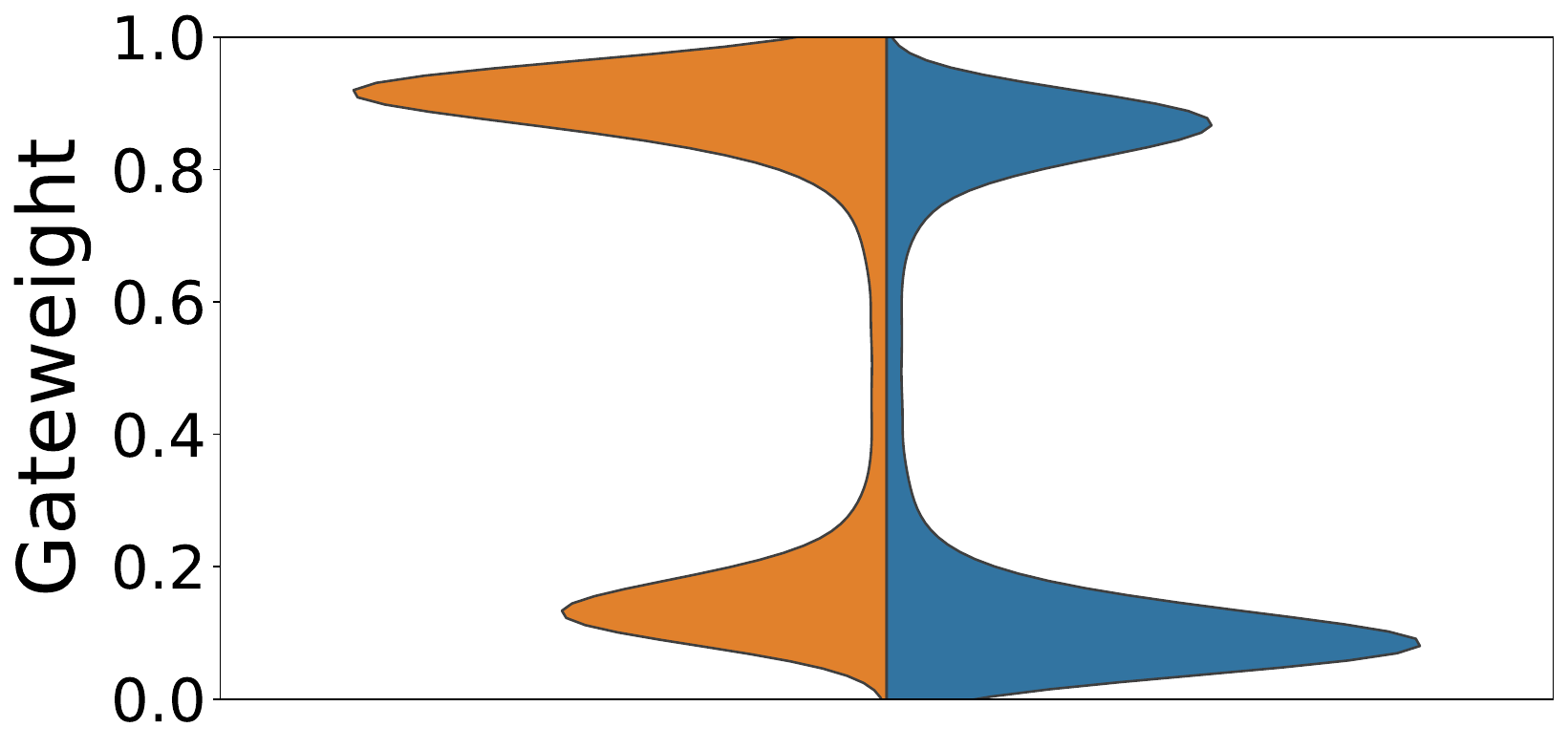}
    \caption*{All subsets}
  \end{subfigure}
\begin{subfigure}[b]{0.49\columnwidth}
    \centering
    \includegraphics[width=\textwidth]{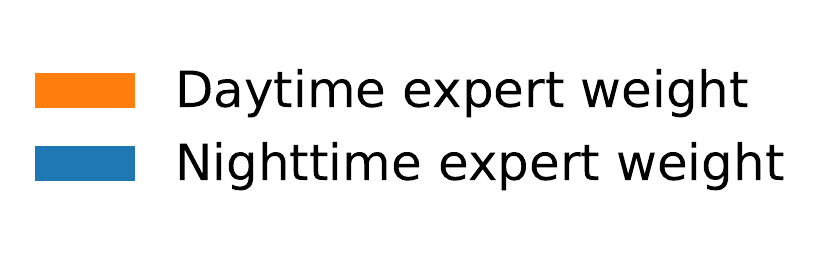}
  \end{subfigure}
  \caption{Distribution of expert weights, predicted by a simple gate in a YOLOv7x-based MoE with samplewise entropy balancing loss and no domain-aware training.}
  \label{fig:violin}
\end{figure}

\textbf{Routing analysis:} The routing analysis provides a global view of expert specialization by examining the distribution of expert weights predicted by the gating network across data subsets. Figure \ref{fig:violin} demonstrates that the gate learns domain-dependent routing behavior, even when no domain-aware training is used. In particular, it consistently routes inputs to the expert trained on the corresponding domain. For daytime images, the gate assigns high weights to the daytime expert and low weights to the nighttime expert, while the opposite behavior is observed for nighttime images, indicating domain-consistent expert weighting. For dawn/dusk images, the routing is skewed toward the daytime expert, suggesting that visual characteristics of this transitional domain are closer to daytime. In contrast, images from the undefined subset exhibit a broad and balanced distribution of gate values across both experts, indicating increased uncertainty and a more even reliance on both models for OOD data.

This observed routing behavior directly enhances interpretability compared to standard ensembles. In ensemble models, predictions are merged through fixed or heuristic postprocessing, making it difficult to attribute a final detection to a specific expert or to analyze expert preference under changing input conditions. In contrast, the MoE explicitly exposes expert contributions through input-dependent gating weights. These weights provide a transparent and quantitative explanation of how much each expert influences the final prediction for a given image. As a result, expert specialization, domain ambiguity, and uncertainty become observable properties of the model rather than implicit effects hidden by postprocessing.

\textbf{Disagreement analysis:} The disagreement analysis focuses on local interpretability by examining how expert predictions differ at the object level. Quantitative results (see Table~\ref{tab:disagreement}) show that expert agreement dominates across all subsets, while a substantial number of BBoxes are predicted by only one expert. Daytime images exhibit a higher number of detections exclusive to the daytime expert, while nighttime images show a corresponding increase in detections exclusive to the nighttime expert. For dawn/dusk, and undefined images, the number of exclusive detections remains high for both experts, indicating increased ambiguity and reduced domain specificity.

\begin{table}[t]
\centering
\caption{Average number of BBoxes per image for each disagreement case.}
\label{tab:disagreement}
\resizebox{1.0\linewidth}{!}{
\begin{tabular}{|r|c|c|c|c|}
    \hline
    \textbf{Data} & \textbf{Full  } & \textbf{Expert } & \textbf{Detection} & \textbf{Detection} \\
    
    \textbf{subset} & \textbf{expert} & \textbf{agreement} & \textbf{only by } & \textbf{only by } \\
     & \textbf{agreement}  & \textbf{on label}  & \textbf{nighttime expert} & \textbf{daytime expert} \\
    \hline
    Daytime     & \textbf{16.12} & 0.29 & 2.97 & \textbf{5.63} \\
    Nighttime   & 14.78 & 0.09 & \textbf{3.29} & 3.90 \\
    Dawn/Dusk   & 15.95 & \textbf{0.21} & 3.00 & 4.69 \\
    Undefined   & 7.81  & 0.07  & 1.49 & 2.28 \\
    All         & 15.54 & 0.20 & 3.07 & 4.72 \\
    \hline
\end{tabular}
}
\end{table}

\begin{figure}[t]
  \centering
  
  \begin{subfigure}[b]{0.75\columnwidth}
    \centering
    \includegraphics[width=\textwidth]{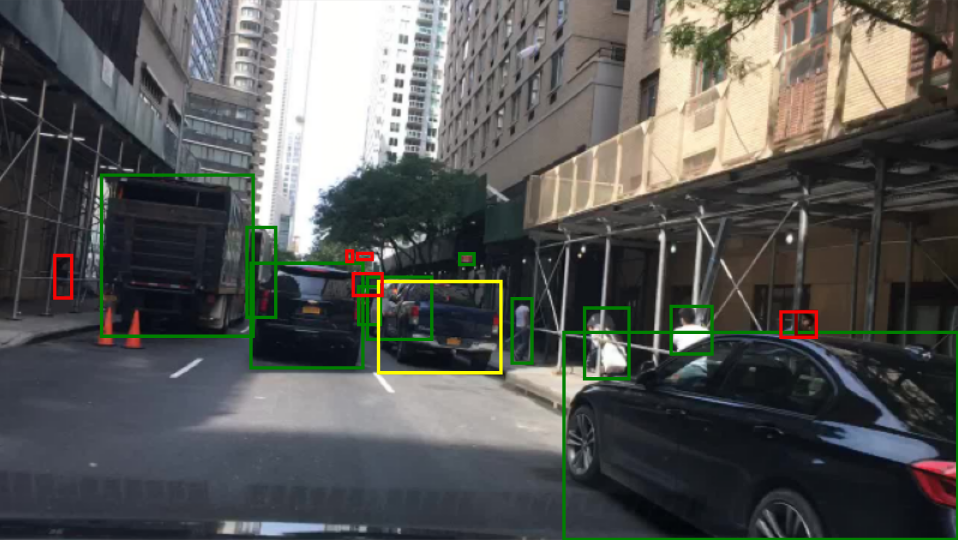}
    \caption*{Daytime subset}
  \end{subfigure}

   \begin{subfigure}[b]{0.75\columnwidth}
    \centering
    \includegraphics[width=\textwidth]{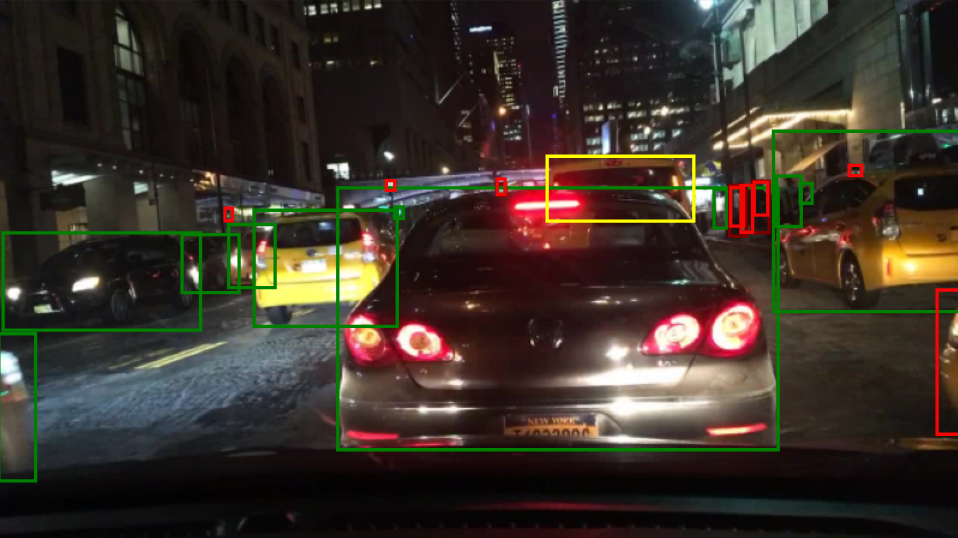}
    \caption*{Nighttime subset}
  \end{subfigure}

  \begin{subfigure}[b]{0.75\columnwidth}
    \centering
    \includegraphics[width=\textwidth]{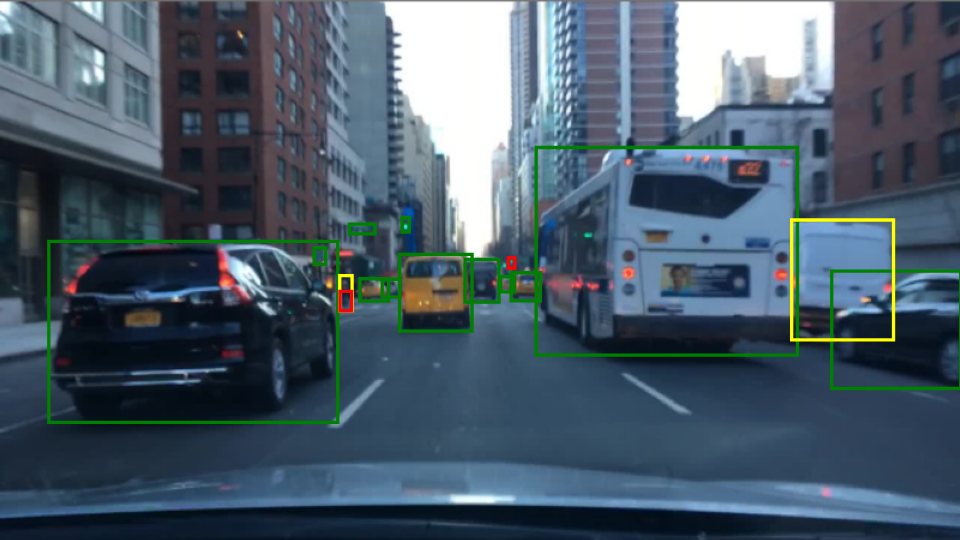}
    \caption*{Dawn/dusk subset}
  \end{subfigure}

  \begin{subfigure}[b]{0.75\columnwidth}
    \centering
    \includegraphics[width=\textwidth]{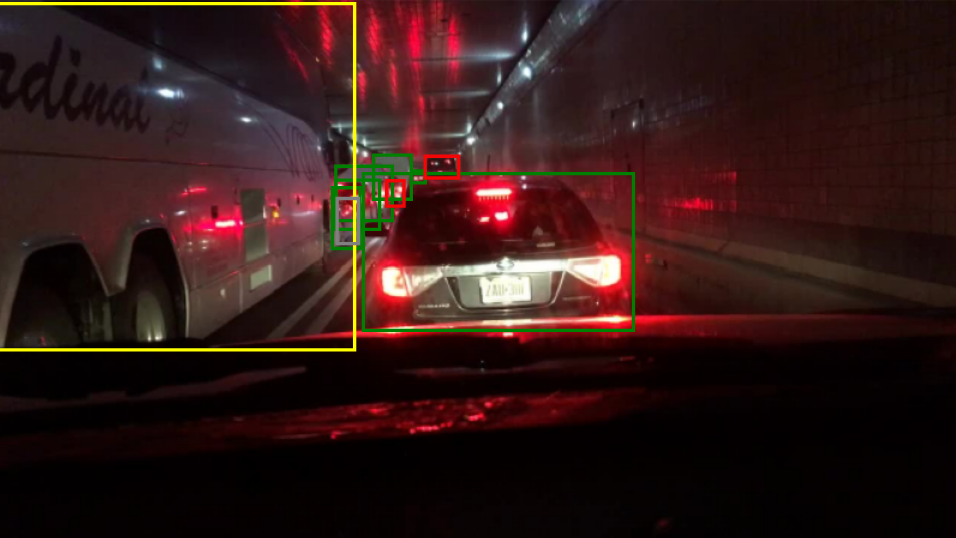}
    \caption*{Undefined subset}
  \end{subfigure}

  \caption{Disagreement analysis helps trace MoE decisions back to expert decisions. BBoxes with full expert agreement are green, BBoxes of expert disagreement on label are yellow, and BBoxes of objects not detected by either expert are red. The results are for the YOLOv7x-based MoE with sample-wise entropy-balancing loss and no domain-aware training.}
  \label{fig:disagreement}
\end{figure}

Qualitative examples (see Figure~\ref{fig:disagreement}) visualize expert agreement and disagreement by indicating for each BBox whether both experts detect the object, disagree on its label, or fail to detect it. Across all subsets, BBoxes with full expert agreement dominate, while a smaller number of objects exhibit label disagreement. Importantly, the examples also show objects that are detected by only one expert (red BBoxes), highlighting missed detections by the other expert. This pattern is visible in all subsets and reflects complementary expert behavior rather than systematic conflicts. By explicitly exposing which expert contributes to each detection and where detections are missing, the MoE enables tracing final predictions back to individual expert decisions. In contrast to standard ensembles, which merge predictions without preserving such attribution, the MoE provides a transparent view of expert agreement, disagreement, and failure cases at the object level.


\section{Conclusion}
In this work, we studied model-level mixtures of experts for object detection and proposed an architecture that combines multiple YOLO-based detectors via a learned gating mechanism applied prior to detection postprocessing. Beyond improving detection performance, the proposed approach explicitly enhances interpretability by exposing input-dependent expert contributions and enabling attribution of detection outcomes to individual experts. Through routing and disagreement analyses, we showed that the model learns meaningful expert specialization across domains and provides insight into how conflicting expert predictions are resolved at both the image and object levels. Experimental results on a semantically split detection task demonstrate consistent improvements over standard ensembling and competitive performance relative to single model baselines, while offering a transparent alternative to heuristic ensemble fusion. The proposed method benefits from modular expert specialization and explicit decision weighting but introduces additional complexity in prediction fusion and relies on meaningful domain separation to realize its advantages. Future work can explore scaling to larger numbers of experts, extending the approach to other detection architectures, and developing more robust gating strategies that explicitly account for uncertainty and sample level ambiguity.

\clearpage
\newpage

\section*{Acknowledgment}

This work was supported by funding from the Topic Engineering Secure Systems of the Helmholtz Association (HGF) and by KASTEL Security Research Labs (46.23.03).

{\small
\bibliographystyle{IEEEtran}
\bibliography{references}

@string{CoRR = "Computing Research Repository (CoRR)"}

@string{CVPR = "Conference on Computer Vision and Pattern Recognition (CVPR)"}

@string{CVPRW = "Conference on Computer Vision and Pattern Recognition (CVPR) - Workshops"}

@string{ECAI = "European Conference on Artificial Intelligence (ECAI)"}

@string{ECCV = "European Conference on Computer Vision (ECCV)"}

@string{ICCV = "International Conference on Computer Vision (ICCV)"}

@string{ICCVW = "International Conference on Computer Vision (ICCV) - Workshops"}

@string{ICLR = "International Conference on Learning Representations (ICLR)"}

@string{IJCNN = "International Joint Conference on Neural Networks (IJCNN)"}

@string{IROS = "International Conference on Intelligent Robots and Systems (IROS)"}

@string{IROSW = "International Conference on Intelligent Robots and Systems (IROS) - Workshops"}

@string{NIPS = "Advances in Neural Information Processing Systems (NIPS)"}

@inproceedings{pavlitskaya2020using,
  author    = {Svetlana Pavlitskaya and
               Christian Hubschneider and
               Michael Weber and
               Ruby Moritz and
               Fabian H{\"{u}}ger and
               Peter Schlicht and
               J. Marius Z{\"{o}}llner},
  title     = {Using Mixture of Expert Models to Gain Insights into Semantic Segmentation},
  booktitle = CVPRW,
  year      = {2020},
}

@incollection{pavlitskaya2022evaluating,
  title={Evaluating Mixture-of-Experts Architectures for Network Aggregation},
  author={Pavlitskaya, Svetlana and Hubschneider, Christian and Weber, Michael},
  booktitle={Deep Neural Networks and Data for Automated Driving: Robustness, Uncertainty Quantification, and Insights Towards Safety},
  year={2022},
}

@article{jacobs1991adaptive,
  author    = {Robert A. Jacobs and
               Michael I. Jordan and
               Steven J. Nowlan and
               Geoffrey E. Hinton},
  title     = {Adaptive Mixtures of Local Experts},
  journal   = {Neural computation},
  volume    = {3},
  number    = {1},
  year      = {1991},
}

@article{shazeer2017outrageously,
  title={Outrageously large neural networks: The sparsely-gated mixture-of-experts layer},
  author={Shazeer, Noam and Mirhoseini, Azalia and Maziarz, Krzysztof and Davis, Andy and Le, Quoc and Hinton, Geoffrey and Dean, Jeff},
  journal={arXiv preprint arXiv:1701.06538},
  year={2017}
}

@inproceedings{pavlitska2025robust,
  title={Robust Experts: the Effect of Adversarial Training on CNNs with Sparse Mixture-of-Experts Layers},
  author={Pavlitska, Svetlana and Fan, Haixi and Ditschuneit, Konstantin and Z{\"o}llner, J Marius},
  booktitle=ICCVW,
  year={2025}
}

@inproceedings{pavlitska2023sparsely,
  title={Sparsely-gated mixture-of-expert layers for cnn interpretability},
  author={Pavlitska, Svetlana and Hubschneider, Christian and Struppek, Lukas and Z{\"o}llner, J Marius},
  booktitle=IJCNN,
  year={2023},
}

@inproceedings{yu2020bdd100k,
  author       = {Fisher Yu and
                  Haofeng Chen and
                  Xin Wang and
                  Wenqi Xian and
                  Yingying Chen and
                  Fangchen Liu and
                  Vashisht Madhavan and
                  Trevor Darrell},
  title        = {{BDD100K:} {A} Diverse Driving Dataset for Heterogeneous Multitask
                  Learning},
  booktitle    = CVPR,
  year         = {2020},
}

@article{xu2021forest,
  title={A forest fire detection system based on ensemble learning},
  author={Xu, Renjie and Lin, Haifeng and Lu, Kangjie and Cao, Lin and Liu, Yunfei},
  journal={Forests},
  year={2021},
}

@inproceedings{wang2023yolov7,
  author       = {Chien{-}Yao Wang and
                  Alexey Bochkovskiy and
                  Hong{-}Yuan Mark Liao},
  title        = {YOLOv7: Trainable Bag-of-Freebies Sets New State-of-the-Art for Real-Time
                  Object Detectors},
  booktitle    = CVPR,
  year         = {2023},
}

@inproceedings{wang2024yolov9,
  author       = {Chien{-}Yao Wang and
                  I{-}Hau Yeh and
                  Hong{-}Yuan Mark Liao},
  title        = {YOLOv9: Learning What You Want to Learn Using Programmable Gradient
                  Information},
  booktitle    = ECCV,
  year         = {2024},
}

@article{wildfire,
    author = {Bahhar, Chayma and Ksibi, Amel and Ayadi, Manel and Jamjoom, Mona M. and Ullah, Zahid and Soufiene, Ben Othman and Sakli, Hedi},
    title = {Wildfire and Smoke Detection Using Staged YOLO Model and Ensemble CNN},
    journal = {Electronics},
    volume = {12},
    year = {2023},
}

@inproceedings{huayhongthong2020incremental,
  title={Incremental Object Detection Using Ensemble Modeling and Deep Transfer Learning},
  author={Huayhongthong, Piyapong and Rerk-u-suk, Siriyakorn and Booddee, Songwit and Padungweang, Praisan and Warasup, Kittipong},
  booktitle={International Conference on Computing and Information Technology},
  year={2020},
}

@article{ultralytics2021yolov5,
  author = {Ultralytics},
  title = {{YOLOv5}: {A} state-of-the-art real-time object detection system},
  year = {2021},
}

@inproceedings{casado2020ensemble,
  author       = {{\'{A}}ngela Casado{-}Garc{\'{\i}}a and
                  J{\'{o}}nathan Heras},
  title        = {Ensemble Methods for Object Detection},
  booktitle    = ECAI,
  year         = {2020},
}

@inproceedings{li2017ensemble,
  title={Ensemble R-FCN for object detection},
  author={Li, Jian and Qian, Jianjun and Zheng, Yuhui},
  booktitle={International Conference on Ubiquitous Information Technologies and Applications},
  year={2017},
}

@inproceedings{peng2018megdet,
  title={Megdet: A large mini-batch object detector},
  author={Peng, Chao and Xiao, Tete and Li, Zeming and Jiang, Yuning and Zhang, Xiangyu and Jia, Kai and Yu, Gang and Sun, Jian},
  booktitle=CVPR,
  year={2018}
}

@article{guo2015deep,
  title={Deep CNN ensemble with data augmentation for object detection},
  author={Guo, Jian and Gould, Stephen},
  journal={arXiv preprint arXiv:1506.07224},
  year={2015}
}

@inproceedings{cai2018cascade,
  title={Cascade r-cnn: Delving into high quality object detection},
  author={Cai, Zhaowei and Vasconcelos, Nuno},
  booktitle=CVPR,
  year={2018}
}

@inproceedings{lee2018ensemble,
  title={An ensemble method of CNN models for object detection},
  author={Lee, Jinsu and Lee, Sang-Kwang and Yang, Seong-Il},
  booktitle={International Conference on Information and Communication Technology Convergence (ICTC)},
  year={2018},
}

@inproceedings{vo2018ensemble,
  title={Ensemble of deep object detectors for page object detection},
  author={Vo, Nguyen D and Nguyen, Khanh and Nguyen, Tam V and Nguyen, Khang},
  booktitle={International Conference on Ubiquitous Information Management and Communication},
  year={2018}
}

@inproceedings{bodla2017soft,
  title={Soft-NMS--improving object detection with one line of code},
  author={Bodla, Navaneeth and Singh, Bharat and Chellappa, Rama and Davis, Larry S},
  booktitle=ICCV,
  year={2017}
}

@inproceedings{zhou2017cad,
  title={Cad: Scale invariant framework for real-time object detection},
  author={Zhou, Huajun and Li, Zechao and Ning, Chengcheng and Tang, Jinhui},
  booktitle=ICCVW,
  year={2017}
}

@inproceedings{redmon2016you,
  title={You only look once: Unified, real-time object detection},
  author={Redmon, Joseph and Divvala, Santosh and Girshick, Ross and Farhadi, Ali},
  booktitle=CVPR,
  year={2016}
}

@article{zhang2024yolo,
  title={A Yolo-based Approach for Fire and Smoke Detection in IoT Surveillance Systems.},
  author={Zhang, Dawei},
  journal={International Journal of Advanced Computer Science \& Applications},
  volume={15},
  number={1},
  year={2024}
}

@article{solovyev2021weighted,
  title={Weighted boxes fusion: Ensembling boxes from different object detection models},
  author={Solovyev, Roman and Wang, Weimin and Gabruseva, Tatiana},
  journal={Image and Vision Computing},
  year={2021},
}

@inproceedings{zhang2019learning,
  title={Learning a mixture of granularity-specific experts for fine-grained categorization},
  author={Zhang, Lianbo and Huang, Shaoli and Liu, Wei and Tao, Dacheng},
  booktitle=ICCV,
  year={2019}
}

@InProceedings{Valada2016a,
  author    = {Valada, Abhinav and Dhall, Ankit and Burgard, Wolfram},
  title     = {{Convoluted Mixture of Deep Experts for Robust Semantic Segmentation}},
  booktitle = IROSW,
  year      = {2016},
}

@inproceedings{mees2016choosing,
  title={Choosing smartly: Adaptive multimodal fusion for object detection in changing environments},
  author={Mees, Oier and Eitel, Andreas and Burgard, Wolfram},
  booktitle=IROS,
  year={2016},
}

@article{oksuz2023mocae,
  author       = {Kemal Oksuz and
                  Selim Kuzucu and
                  Tom Joy and
                  Puneet K. Dokania},
  title        = {MoCaE: Mixture of Calibrated Experts Significantly Improves Object
                  Detection},
  journal      = {Trans. Mach. Learn. Res.},
  year         = {2024},
}

@inproceedings{lin2014microsoft,
  title={Microsoft coco: Common objects in context},
  author={Lin, Tsung-Yi and Maire, Michael and Belongie, Serge and Hays, James and Perona, Pietro and Ramanan, Deva and Doll{\'a}r, Piotr and Zitnick, C Lawrence},
  booktitle=ECCV,
  pages={740--755},
  year={2014},
  organization={Springer}
}

@article{fedus2021switch,
  author    = {William Fedus and
               Barret Zoph and
               Noam Shazeer},
  title     = {Switch Transformers: Scaling to Trillion Parameter Models with Simple
               and Efficient Sparsity},
  journal   = {CoRR},
  volume    = {abs/2101.03961},
  year      = {2021},
}

@inproceedings{lepikhin2021gshard,
    author = { HyoukJoong Lee and
               Yuanzhong Xu and
               Dehao Chen and
               Orhan Firat and
               Yanping Huang and
               Maxim Krikun and
               Noam Shazeer and
               Zhifeng Chen},
  title     = {GShard: Scaling Giant Models with Conditional Computation and Automatic
               Sharding},
  booktitle = ICLR,
  year      = {2021},
}

@inproceedings{riquelme2021scaling,
  author    = {Carlos Riquelme and
               Joan Puigcerver and
               Basil Mustafa and
               Maxim Neumann and
               Rodolphe Jenatton and
               Andr{\'{e}} Susano Pinto and
               Daniel Keysers and
               Neil Houlsby},
  title     = {Scaling Vision with Sparse Mixture of Experts},
  booktitle = NIPS,
  year      = {2021},
}

@article{dai2024deepseekmoe,
  title={Deepseekmoe: Towards ultimate expert specialization in mixture-of-experts language models},
  author={Dai, Damai and Deng, Chengqi and Zhao, Chenggang and Xu, RX and Gao, Huazuo and Chen, Deli and Li, Jiashi and Zeng, Wangding and Yu, Xingkai and Wu, Yu and others},
  journal={arXiv preprint arXiv:2401.06066},
  year={2024}
}

@inproceedings{ahmed2016network,
  title={Network of experts for large-scale image categorization},
  author={Ahmed, Karim and Baig, Mohammad Haris and Torresani, Lorenzo},
  booktitle=ECCV,
  year={2016},
  organization={Springer}
}

@article{rudin2019stop,
  author       = {Cynthia Rudin},
  title        = {Stop explaining black box machine learning models for high stakes
                  decisions and use interpretable models instead},
  journal      = {Nat. Mach. Intell.},
  year         = {2019},
}

@article{doshi2017towards,
  title={Towards a rigorous science of interpretable machine learning},
  author={Doshi-Velez, Finale and Kim, Been},
  journal={arXiv preprint arXiv:1702.08608},
  year={2017}
}

@article{yang2026fg,
  title={FG-MoE: Heterogeneous mixture of experts model for fine-grained visual classification},
  author={Yang, Songming and Wen, Jing and Fang, Bin},
  journal={Pattern Recognition},
  pages={113050},
  year={2026},
  publisher={Elsevier}
}
}

\end{document}